\documentclass[10pt,twocolumn,letterpaper]{article}
\usepackage{booktabs}
\usepackage{makecell}
\usepackage{wacv}              

\definecolor{wacvblue}{rgb}{0.21,0.49,0.74}
\usepackage[pagebackref,breaklinks,colorlinks,allcolors=wacvblue]{hyperref}
\usepackage{booktabs}

\def\wacvPaperID{*****} 
\def\confName{WACV}
\def\confYear{2026}

\usepackage{algorithm}
\usepackage{algorithmic}

\title{
Motion Focus Recognition in Fast-Moving Egocentric Video
}
\vspace{-0.5cm}
\author{Si-En Hong$^1$, James Tribble$^1$, Alexander Lake$^1$, Hao Wang$^1$, Chaoyi Zhou$^1$, Ashish Bastola$^1$,\\ Siyu Huang$^1$, Eisa Chaudhary$^2$, Brian Canada$^2$, Ismahan Arslan-Ari$^3$, and Abolfazl Razi$^1$\thanks{Corresponding author.}\\
$^1$Clemson University, SC, USA\\
$^2$University of South Carolina Beaufort, SC, USA\\
$^3$University of South Carolina, SC, USA\\
{\tt\small \{sienh, jjtribb, lake8, hao9, chaoyiz, abastol, siyuh, arazi\}@clemson.edu
}\\
{\tt\small eisa@email.uscb.edu, bcanada@uscb.edu, arslanai@mailbox.sc.edu}
}

\vspace{-0.5cm}

\begin{document}
\maketitle

\begin{abstract}
\vspace{-0.5cm}

From Vision–Language–Action (VLA) systems to robotics, existing egocentric datasets primarily focus on action recognition tasks, while largely overlooking the inherent role of motion analysis in sports and other fast-movement scenarios. 
To bridge this gap, we propose a real-time motion focus recognition method that estimates the subject’s locomotion intention from any egocentric video. We leverages the foundation model for camera pose estimation and introduces system-level optimizations to enable efficient and scalable inference. 
Evaluated on a collected egocentric action dataset, our method achieves real-time performance with manageable memory consumption through a sliding batch inference strategy. This work makes motion-centric analysis practical for edge deployment and offers a complementary perspective to existing egocentric studies on sports and fast-movement activities.\footnote{Research reported in this publication was supported in part by the NSF and SC EPSCoR Program under award number (NSF Award \# OIA-2242812 and SC EPSCoR 26-CRP03). The views, perspective, and content do not necessarily represent the official views of the SC EPSCoR Program nor those of the NSF.}
Our code and dataset are available on: 
\url{https://arazi2.github.io/aisends.github.io/project/VisionGPT}

\vspace{-0.5cm}
\end{abstract}
\section{Introduction}
\label{sec:intro}


In fast human movement, egocentric perception is fundamentally more challenging than vision tasks developed for autonomous vehicles or rigid robotic platforms \cite{WaymoDataset}. 
Unlike vehicle-mounted cameras operating under stable kinematic constraints, egocentric video is shaped by frequent head movements, body sway, and intentional gaze shifts, leading to unstable viewpoints and noisy motion feedback \cite{PedestrianNavigationChallenges}. 
In such settings, common assumptions, such as image-center relevance and consistent camera orientation, break down, making reliable real-time ego-motion estimation particularly challenging \cite{liu2020forecasting}.

Most existing attention and relevance estimation methods for egocentric vision rely primarily on static or appearance-based saliency to identify important regions and objects~\cite{qiao2024saliency}. 
While effective in relatively stable settings, such approaches can introduce substantial bias in high-motion environments, as visually salient but motion-inconsistent objects often appear only transiently and disappear within a few frames. 
Consequently, static saliency fails to capture the temporally persistent information that is most informative for understanding ego-motion and short-term movement intent \cite{yang2024agent}.

\begin{figure}[htbp]
    \centering
    \centerline{\includegraphics[width=1\linewidth]{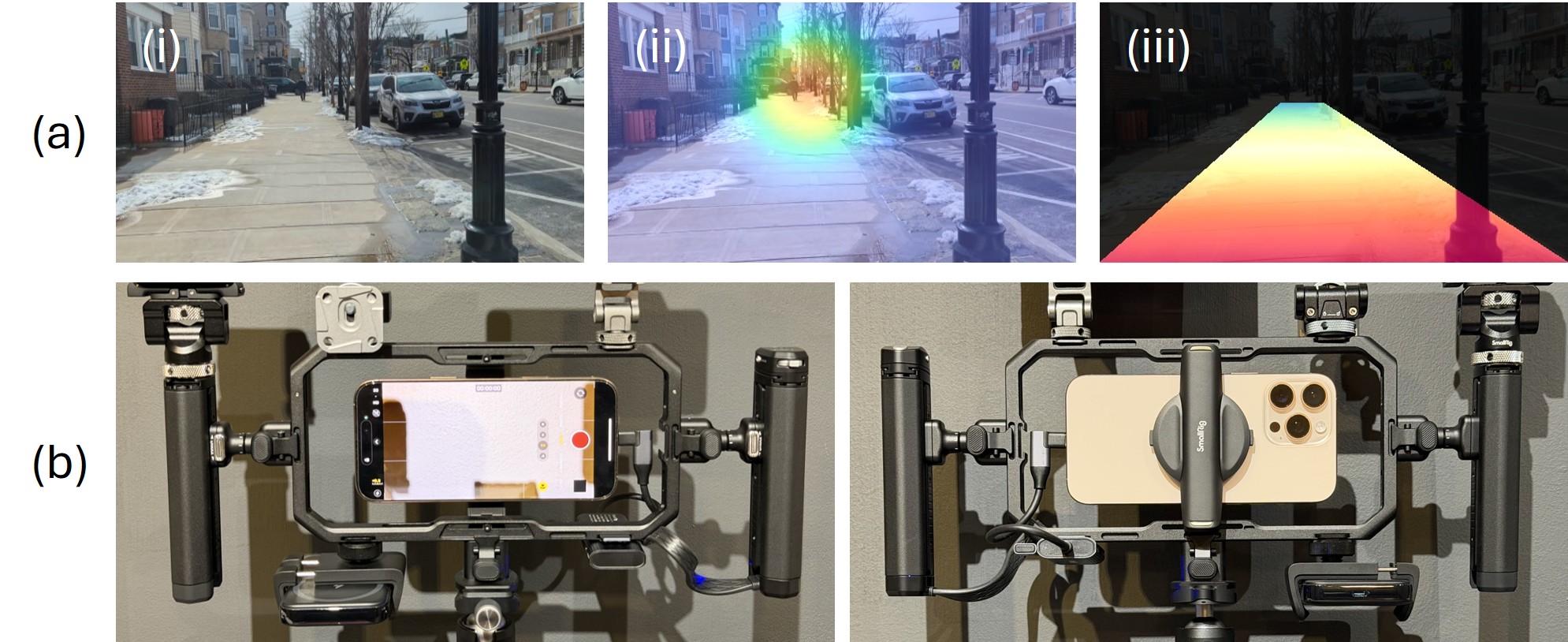}}
    \caption{(a) Selected frame from the dataset collected during the winter season. (i) is the raw city street image, (ii) is the motion-focus map, and (iii) is the motion-guided depth map. (b) is the example camera system for data capturing.}
    \label{fig:cover}
    \vspace{-0.25cm}
\end{figure}

\begin{figure*}[htbp]
    \centering
    \centerline{\includegraphics[width=1\textwidth]{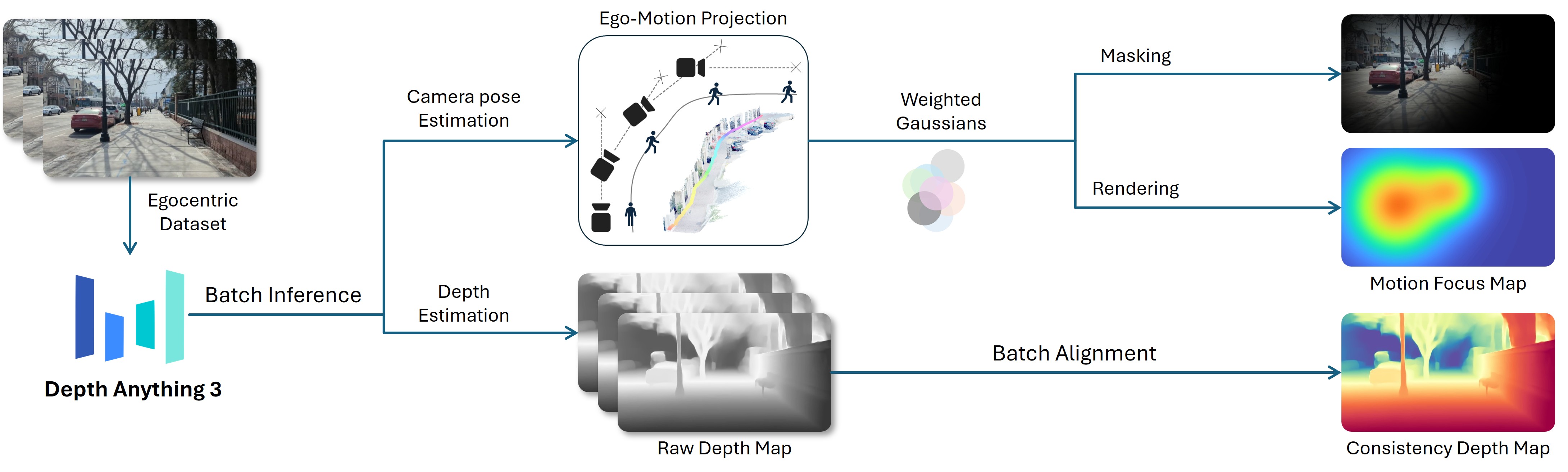}}
    \caption{Framework for motion focus prediction and projection.}
    \label{fig:frame}
    
\vspace{-0.5cm}
\end{figure*}

To address this limitation, we propose a real-time motion focus recognition method that estimates human locomotion trends from egocentric video while being robust to irregular camera movement noise induced by natural head and body movements. 
Our approach is grounded in physical priors of camera pose dynamics and does not require additional training, enabling flexible motion trend estimation under highly dynamic viewpoints. 
Furthermore, the proposed method can be embedded into arbitrary camera pose estimation frameworks with minimal computational cost, supporting real-time deployment on resource-constrained systems in fast-movement settings.

Specifically, we adopt the camera pose prediction from an existing foundation model and apply deep systematic optimization for real-time egocentric motion recognition~\cite{DA3}. 
To overcome the shortage of existing egocentric datasets in fast-movement scenarios, we collect a lightweight egocentric action dataset and incorporate real camera intrinsics as physical constraints for error compensation, enabling more stable motion focus recognition under real-world capture conditions. 
Experimental results show that the predicted motion focus highlights the videographer’s locomotion trend.


\vspace{-0.5cm}
\section{Method}
\label{sec:method}

\subsection{Data Collection}

We collect an egocentric action video dataset through in-person acquisition to capture realistic human locomotion under dynamic environments. 
All videos are recorded using a single monocular handheld mobile phone carried by a walking subject, resulting in camera motion that reflects natural translation, head turns, and body-induced perturbations. 

\begin{table}[htbp]  
    \centering      
        \resizebox{1\linewidth}{!}{
    \begin{tabular}{c|c|c|c|c|c} 
        \toprule      
        Scene  &Clips  &Resolution&FPS&Total length& Total Frames\\\midrule
 Suburban&42  &960x544&30&37 mins& 65934\\
 City &56  &1920x1080&30&79 mins&  5464\\
 Town&36  &3840x2160&30&24 mins&  21528\\
        Campus&72  &1920x1080&60&75 mins&  269550\\ \bottomrule
    \end{tabular}}
    \caption{Collected egocentric video clips for motion analysis tasks.}
        \label{tab:data}
\end{table}


Data is collected during winter conditions across diverse and challenging environments, including outdoor sidewalks and pedestrian walkways with uneven surfaces and icy roads, as shown in Figure~\ref{fig:cover}.
The dataset further includes multiple locomotion patterns, such as walking, scooter riding, and biking, together with motion variations including turning, slowing down, and brief stopping, producing rich ego-motion dynamics.
In total, the dataset contains action video clips captured at varying resolutions and frame rates, as summarized in Table~\ref{tab:data}.

\subsection{System Optimization for Real-Time Inference}
\label{subsec:depth}

\begin{figure}[htbp]
    \centering
    \centerline{\includegraphics[width=1\linewidth]{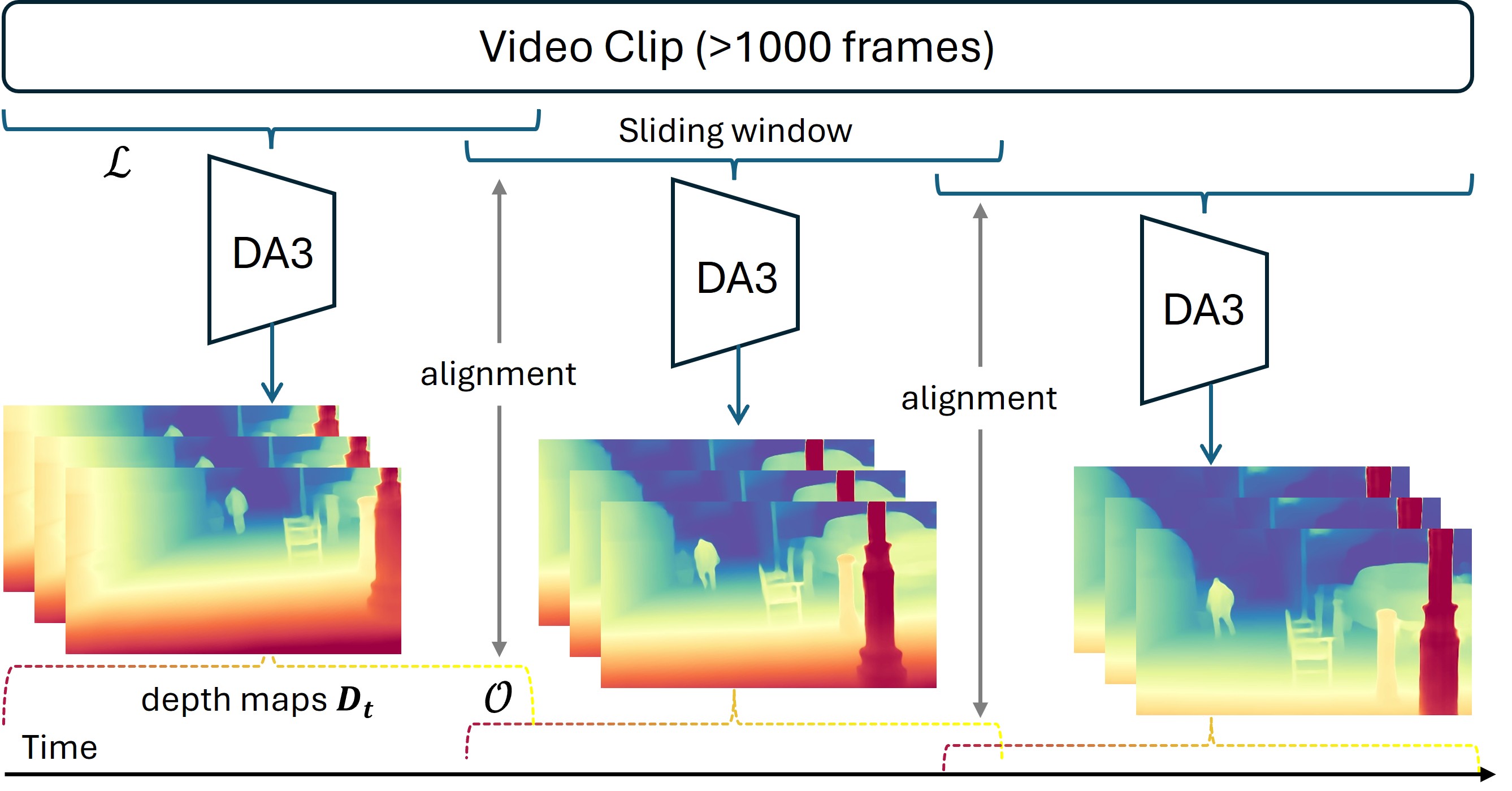}}
    \caption{Depth alignment in the real-time inference. We applied a sliding window for consecutive inference in real time.}
    \label{fig:depth}
\end{figure}

Since Depth Anything 3 is primarily designed for offline inference in industrial applications \cite{DA3}, processing an entire long video in one pass quickly exceeds GPU memory. During our testing on a 24 GB desktop GPU, sequences exceeding 1000 frames fail due to running out of memory. 
To avoid this, we split the video into overlapping batches, perform inference per batch, and align them temporally. This reduces peak memory usage while preserving a complete, consistent sequence of depth maps and camera poses, as shown in Figure \ref{fig:depth}.

Specifically, to handle extended video streams $\{\mathbf{I}_t\}$ under memory constraints while maintaining real-time performance, we adopt a sliding-window inference strategy. The continuous stream is partitioned into overlapping segments (batches) of size $\mathcal{L}$, where consecutive windows share $\mathcal{O}$ frames to ensure temporal context, as shown in Figure \ref{fig:depth}.

For each batch, the model predicts depth maps $\{\mathbf{D}_t\}$ and camera extrinsics $\{\mathbf{T}_t^{w \rightarrow c}\}$ in a local coordinate frame. To resolve the coordinate ambiguity between consecutive batches, we perform an incremental anchoring process. Let $t_{anchor}$ denote the transition frame shared by the current and previous windows. We compute a rigid transformation to align the current batch's local origin to the terminal pose of the previous trajectory, effectively stitching the segments into a unified world coordinate system. 

Specifically, to maintain the physical consistency of the motion, we lock the gravity-aligned rotation (pitch and roll) and propagate only the horizontal yaw and translation across batch boundaries. This incremental anchoring avoids the global drift typical of large-scale Sim(3) optimization while preserving the predictive power of the foundation model for long-range navigation \cite{DA3}. Following the alignment, only the newly processed, non-overlapping frames are emitted, ensuring a continuous and temporally coherent flow of depth and pose estimates for downstream motion analysis.

\begin{table}[htbp]
    \centering
        \resizebox{0.9\linewidth}{!}{
        \begin{tabular}{c|cccc}\toprule
           Model&Resolution&Batch Size&  FPS&GPU Mem.\\ \midrule
           DA3 Base&504&60&  31.53&8113 MB\\
 DA3 Base& 384& 60& 34.01&5556 MB\\
 DA3 Base& 384& 30& 28.65&4501 MB\\\midrule
 DA3 Small& 504& 60& 32.95&4609 MB\\
 DA3 Small& 384& 60& 38.44&3437 MB\\
           DA3 Small&504&30&  35.87&3677 MB\\ 
 DA3 Small& 384& 30& 32.32&3064 MB\\\bottomrule
    \end{tabular}}
    \caption{Performance of DA3.}
    \label{tab:da3}
\end{table}


We evaluate the batch inference strategy under different configurations to assess its suitability for real-time deployment. As shown in Table~\ref{tab:da3}, the DA3 base model achieves 31.53 FPS with approximately 8\,GB GPU memory at a batch size of 60, and further improves to 34.01 FPS with reduced memory usage when the resolution is lowered to 384\,px. While smaller batch sizes reduce memory consumption, they also lead to decreased throughput due to increased I/O and scheduling overhead. By replacing the base model with the DA3 small variant, our framework maintains stable real-time performance above 30 FPS with GPU memory usage below 5\,GB, enabling practical deployment on consumer-grade hardware.

\subsection{Motion Focus Recognition}
\label{sec:motion_focus}

To enable motion-aware prioritization in fast movements, we introduce a lightweight Motion Focus Recognition module that estimates the impact region induced by the user’s locomotion. 
The core intuition is that the acceleration vector induced by camera translation, when projected onto the image plane, provides a physically grounded focus of attention indicating where near-field interactions or potential collisions are most likely to occur~\cite{fukuchi2009focus}.

Let $\mathbf{T}_{t}^{w \rightarrow c} \in \mathbb{R}^{4 \times 4}$ denote the camera pose at time $t$, expressed as a world-to-camera transformation. We first recover the camera positions in the world coordinate system by inverting the poses:
\begin{equation}
\mathbf{T}_{t}^{c \rightarrow w} = \left(\mathbf{T}_{t}^{w \rightarrow c}\right)^{-1}, \quad
\mathbf{p}_t = \mathbf{T}_{t}^{c \rightarrow w}[1{:}3,\,4].
\end{equation}

To characterize the dynamic trend of the trajectory, we compute the discrete acceleration vector in world space, $\mathbf{a}_w$, by considering three consecutive pose samples:
\begin{equation}
\mathbf{v}_t = \mathbf{p}_t - \mathbf{p}_{t-1}, \quad \mathbf{v}_{t-1} = \mathbf{p}_{t-1} - \mathbf{p}_{t-2}
\end{equation}
\begin{equation}
\mathbf{a}_w = \mathbf{v}_t - \mathbf{v}_{t-1} = \mathbf{p}_t - 2\mathbf{p}_{t-1} + \mathbf{p}_{t-2}.
\end{equation}

To align this motion trend with the user's current egocentric perspective, the acceleration vector is transformed from the world frame into the local camera coordinate frame:
\begin{equation}
\mathbf{a}_c = \mathbf{R}_{t}^{w \rightarrow c} \, \mathbf{a}_w,
\end{equation}
where $\mathbf{R}_{t}^{w \rightarrow c}$ is the rotation matrix of $\mathbf{T}_{t}^{w \rightarrow c}$. The local vector $\mathbf{a}_c = (a_x, a_y, a_z)^\top$ encapsulates the directional change of motion relative to the optical axis.

Assuming a pinhole camera model with an intrinsic matrix
\begin{equation}
\mathbf{K} =
\begin{bmatrix}
f_x & 0 & c_x \\
0 & f_y & c_y \\
0 & 0 & 1
\end{bmatrix},
\end{equation}
We project the motion direction onto the image plane to obtain an estimated impact point:
\begin{equation}
u_{acc} = c_x + f_x \frac{a_x}{a_z}, \quad v_{acc} = c_y + f_y \frac{a_y}{a_z}.
\end{equation}

\begin{figure}[htbp]
    \centering
    \centerline{\includegraphics[width=1\linewidth]{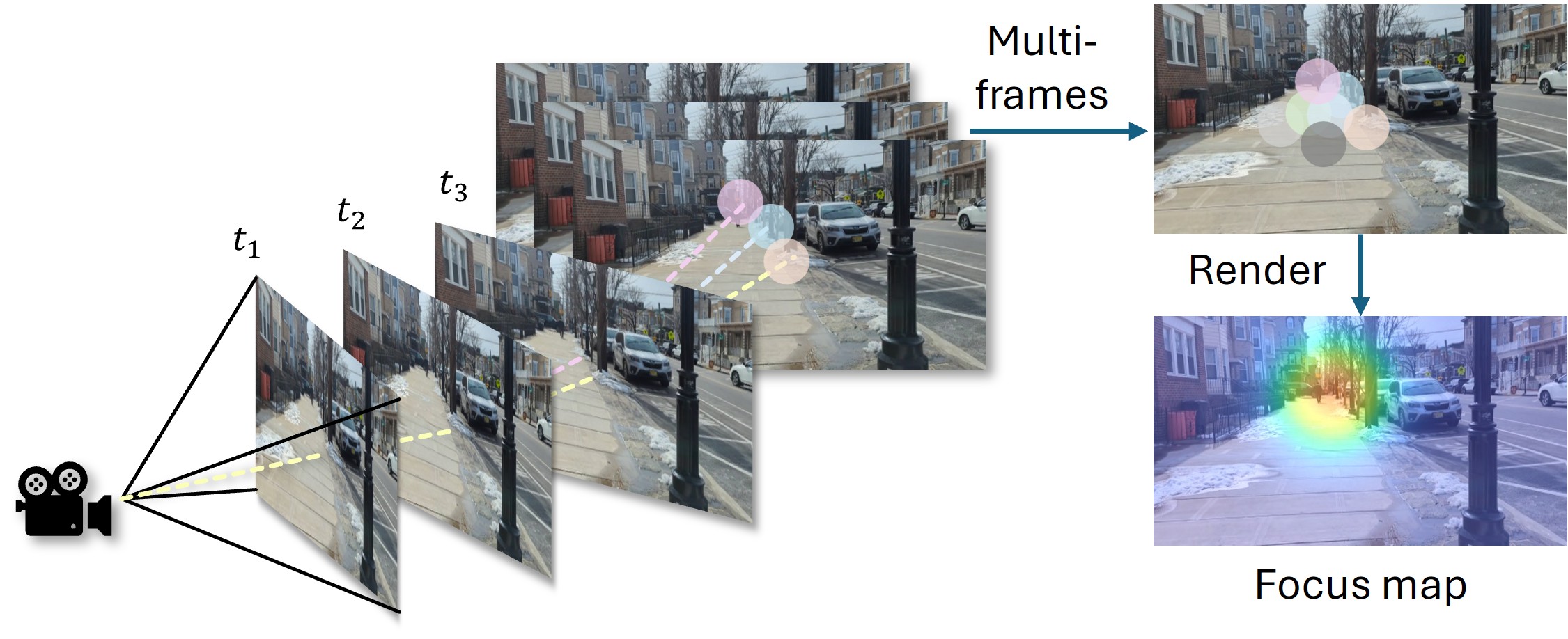}}
    \caption{Motion focus calculation process. We first calculate the camera center and project the acceleration vector to the $n$-th frame. Then we aggregate $N$ camera centers and apply Gaussian kernels for rendering. The final results represent the camera movement trends, which aligned with the defined motion focus.}
    \label{fig:motion}
\end{figure}

The coordinate $(u_{acc}, v_{acc})$ serves as the center of a motion-induced saliency map. Unlike velocity-based projection, this acceleration-based focus effectively anticipates the user’s intent to turn or decelerate, as shown in Figure \ref{fig:motion}.

\begin{figure*}[htbp]
    \centering
    \centerline{\includegraphics[width=1\textwidth]{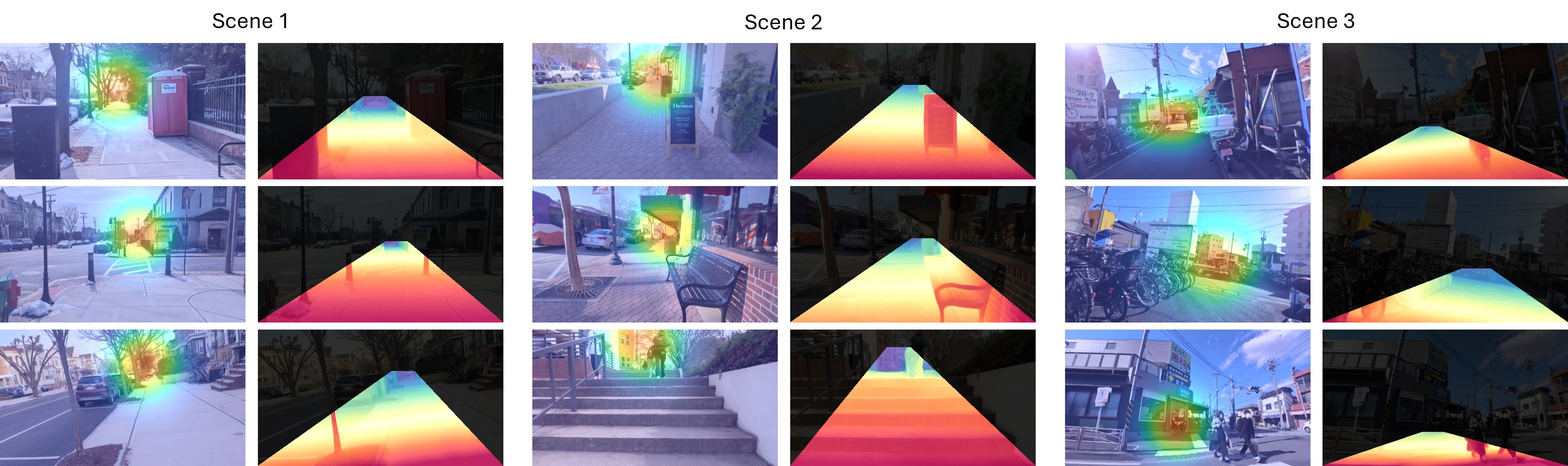}}
    \caption{Visualization of motion focus. Left: raw image with predicted motion focus map; Right: depth map guided by motion direction.}
    \label{fig:demo}
    
\vspace{-0.5cm}
\end{figure*}

As shown in Figure \ref{fig:motion}, we then construct a motion focus map by aggregating $N$ motion points (e.g., $N=15$) in the image plane.
Each motion point contributes a Gaussian kernel centered at its projected pixel location, with the kernel spread modulated by the corresponding motion magnitude to reflect its spatial influence.
The final focus map is obtained by summing all kernels and normalizing the result:
\begin{equation}
M(u,v) = \frac{1}{Z} \sum_{i=1}^{N} \exp\!\left(-\frac{(u-u_i)^2 + (v-v_i)^2}{2(\sigma s_i)^2}\right),
\end{equation}
where $(u_i, v_i)$ denotes the $i$-th motion point, $s_i$ its motion magnitude, $\sigma$ a scale factor, and $Z$ a normalization constant.

The resulting motion focus allowing the system to prioritize recognition in regions that align with the predicted future path.



\subsection{Qualitative Analysis for Motion Focus}
\label{subsec:qualitative}

We conduct qualitative analysis on 30 video clips and select 3 representative scenes. Visual inspection reveals that the proposed motion focus mechanism operates robustly across diverse navigation scenarios, consistently highlighting motion-relevant regions.

As illustrated in Figure~\ref{fig:demo}, the proposed motion focus consistently highlights motion-relevant regions across diverse egocentric scenarios. 
In  Scene 1 (scooter), static but semantically salient objects such as parked vehicles are suppressed, while nearby obstacles along the walking path (e.g., poles, trees, and trash bins) are emphasized, demonstrating effective filtering of action-irrelevant visual clutter. 
In Scene 2 (walking), the predicted motion center adapts to vertical locomotion, shifting toward the ascending direction rather than remaining fixed at the image center. 
Furthermore, during Scene 3 (biking), when head orientation and body movement direction diverge, motion focus remains aligned with the torso-driven locomotion trend, prioritizing areas along the actual movement path.



\section{Conclusion}
\label{sec:conclusion}


This work shows that motion–driven analysis can serve as a lightweight and practical foundation for real-time egocentric perception, complementing existing action- and semantics-driven approaches in sports and fast-movement scenarios.
More broadly, our findings suggest that grounding egocentric perception in physically significant motion is essential for understanding sports, robotics, and other dynamic embodied activities. By enabling lightweight, real-time ego-motion reasoning, this work takes a step toward more reliable and human-centered perception systems that can support safety, performance analysis, and assistive feedback in challenging real-world environments.



{
    \small
    \bibliographystyle{ieeenat_fullname}
    \bibliography{main}
}

\end{document}